\documentclass[conference]{IEEEtran}

\IEEEoverridecommandlockouts

\usepackage[pdftex]{graphicx}
\graphicspath{{./pdf/}{./jpeg/}}
\DeclareGraphicsExtensions{_compressed.pdf,.jpeg,.png}

\usepackage[cmex10]{amsmath}
\usepackage{mathabx}
\usepackage{balance}
\usepackage{cite}
\usepackage{array}
\usepackage{algorithm}
\usepackage{algpseudocode}
\usepackage{mdwmath}
\usepackage{mdwtab}
\usepackage{eqparbox}
\usepackage{url}
\usepackage{mwe}
\usepackage{amsmath}
\usepackage{mathtools}
\usepackage{subfigure}
\usepackage{graphicx}
\usepackage{paralist}
\usepackage{cite}
\begin{document}

\title{\LARGE Wedge-Jamming Mechanism for String Configuration}

\author{
Taro Tako$^{1,*}$, Takanori Jin$^{1}$, Hiroshi Kaminaga$^{2}$, and Taisuke Kobayashi$^{1}$
\thanks{$^{1}$T. Tako, T. Jin, and T. Kobayashi are with the National Institute of Informatics (NII) and with The Graduate University for Advanced Studies (SOKENDAI), 2-1-2 Hitotsubashi, Chiyoda-ku, Tokyo, 101-8430, Japan}%
\thanks{$^{2}$ H. Kaminaga is with the CNRS-AIST JRL (Joint Robotics Laboratory), IRL 3218, National Institute of Advanced Industrial Science and Technology (AIST), 1-1-1 Umezono, Tsukuba, Ibaraki 305-8568, Japan}%
\thanks{$^{*}$ Corresponding author: tako@nii.ac.jp}
}

\maketitle

\begin{abstract}
Variable stiffness mechanisms that achieve both flexibility and rigidity are important technologies for robotic systems.
Recently, wire-based jamming mechanisms have attracted significant attention.
In the wire-based jamming mechanisms, modules are chained by a wire, and the overall stiffness is controlled by changing the friction between the modules through wire tension.
These mechanisms have been widely studied due to their simple structure and controllability.
However, conventional wire-based jamming mechanisms mainly focus on holding torque along the joint axis, and the resistance against disturbances in other directions has not been sufficiently explored.
In this study, we propose a wedge-jamming mechanism that introduces wedge-shaped contact surfaces into the modules of wire-based jamming structures.
By changing the wedge angle, the holding characteristics against disturbances in different directions can be adjusted.
Experimental results show that an out-wedge configuration improves resistance to vertical disturbances, while an in-wedge configuration enhances resistance to torsional disturbances.
These results demonstrate that the anisotropic holding performance of wire-based jamming mechanisms depend on wedge geometry.
\end{abstract}



\section{Introduction}
\label{sec:introduction}
In humans, force transmission is achieved through the musculoskeletal system, consisting of muscles, tendons, and the skeletal structure.
This system enables a balance between flexibility and stiffness, which are inherently conflicting properties.
Flexibility allows safe interaction with external forces and impacts, whereas stiffness is essential for efficient force transmission and accurate positioning\cite{Hogan1985impedance}.

Replicating this balance in robotic systems requires mechanisms capable of adapting their stiffness according to task demands.
Therefore, variable stiffness mechanisms play a crucial role in robotic systems\cite{lin2024variable}.

Among various variable stiffness mechanisms, wire-based linear structures have attracted attention due to their simple mechanical design and high controllability.
Such mechanisms have been widely applied to robotic hands and grippers\cite{tadakuma2020fire}\cite{hara2025juzu}\cite{lu2025kangaroo}.
In these systems, multiple modules are chained by a wire.
By changing the wire tension, the frictional forces between adjacent modules are modified, resulting in variable stiffness characteristics accordingly.

However, the holding performance of wire-based variable stiffness mechanisms remains limited, particularly under high external loads.
To address this issue, this study proposes a novel module design based on a wedge-shaped structure for variable stiffness mechanisms using linear configurations.
The proposed design enhances holding performance by effectively increasing frictional forces under load.
The effectiveness of the proposed mechanism is evaluated through analytical modeling and experimental validation.

\section{Related work}
\label{sec:related_work}
In soft robotics, various variable stiffness mechanisms have been proposed.
These studies have realized methods for controllably adjusting stiffness.
Existing variable stiffness mechanisms can be broadly classified into friction-based approaches and material property-based approaches.

Granular jamming\cite{cates1998jamming,roux2000geometric,brown2010universal} is a representative example of friction-based jamming.
In this method, granular materials are enclosed in a flexible chamber, and stiffness is increased by applying negative pressure.
The stiffness can be controlled by adjusting the applied pressure.
However, since force generation depends on the inter-particle distance, the shape of the flexible chamber is constrained.
In addition, a pneumatic system is required, which makes integration into robots difficult.

Magneto-rheological jamming\cite{majidi2010tunable,pettersson2010design} and electro-rheological fluid jamming\cite{sadeghi2012innovative} are material property-based approaches.
These methods enclose functional fluids in a flexible chamber, whose viscoelasticity properties change in response to externally applied magnetic or electric fields.
However, magneto-rheological fluids require non-magnetic objects for grasping.
In addition, electro-rheological fluids exhibit low shear strength, making it difficult to generate sufficient force even with high-voltage power supplies.
Moreover, these materials pose environmental concerns.

\begin{figure}[tb]
    \centering
    \includegraphics[width=1.0\linewidth]{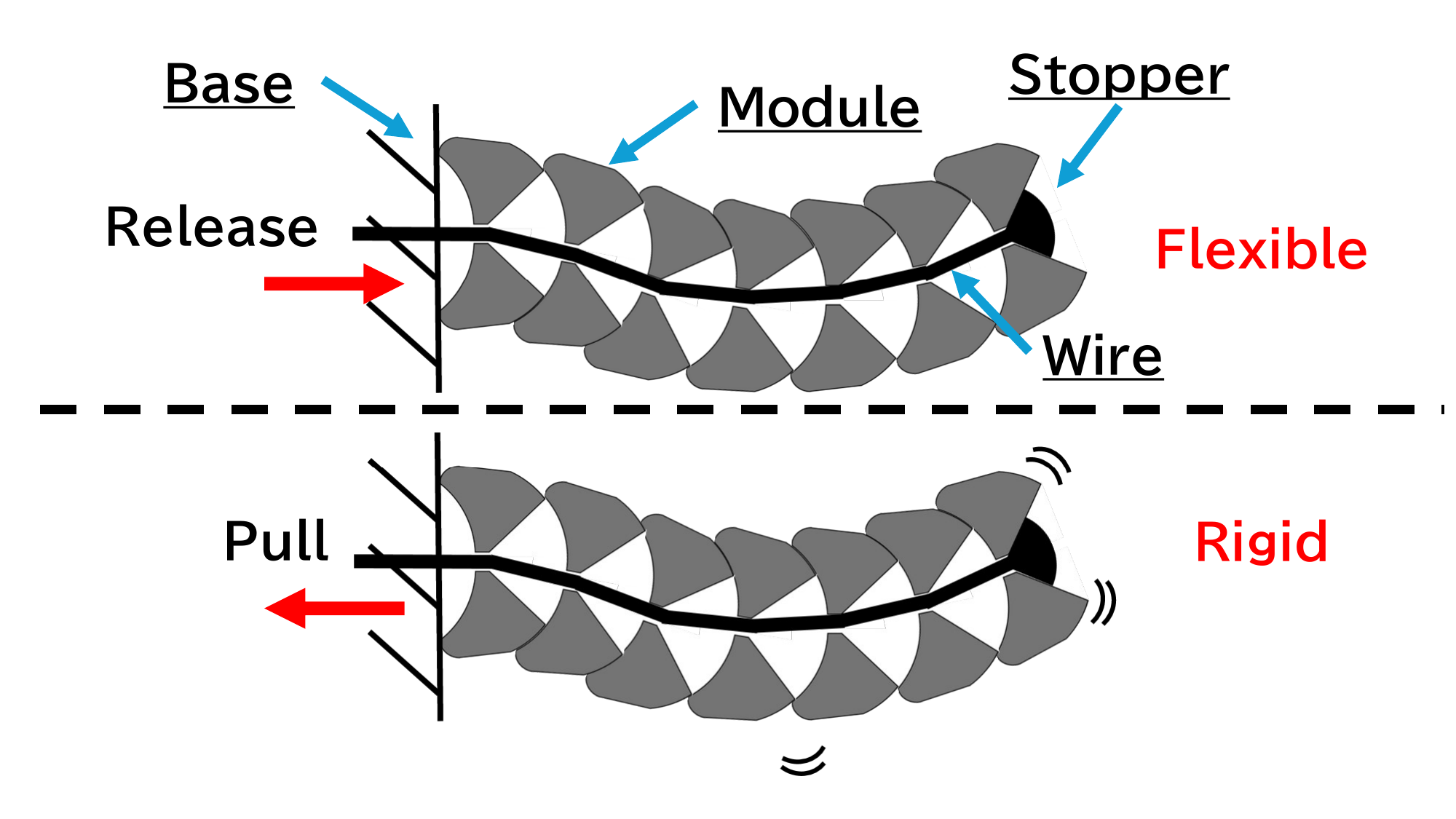}
    \caption{Principle of wire-based jamming mechanism}
    \label{fig:principle_of_wire_jamming}
\end{figure}

This study focuses on wire-based jamming mechanisms\cite{jiang2019chain,mukaide2020radial,tadakuma2020fire,hara2025juzu,lu2025kangaroo} employing linear structures (see Fig.~\ref{fig:principle_of_wire_jamming}).
In the wire-based jamming mechanisms, modules are connected in series by a wire, and the overall stiffness is adjusted by changing the friction between adjacent modules through wire tension.
Since the wire tension can control with high-precision, responsive electromotor, these mechanisms offer high controllability of stiffness.
Originally, they have been utilized in surgical device (e.g., a flexible tendon-controlled device for endoscopy \cite{cates1998jamming} and highly articulated robotic probe for minimal invasive surgery\cite{sturges1993flexible}).
Furthermore, the wire-based jamming mechanisms have recently been applied to various systems, fire-resistant grippers\cite{tadakuma2020fire}, Juzu-type grippers\cite{hara2025juzu} and rescue robot grippers\cite{lu2025kangaroo}.
However, there exists a structural limitation on the holding torque against external forces and torsional loads, because stiffness depends on both inter-module friction and wire tension.
To address this issue, several studies have tackled to enhance friction with modifications to the contact surface geometry of the modules, such as radial-layer-type jamming\cite{mukaide2020radial}.

In the most previously proposed wire-based jamming mechanisms, the contact surfaces between adjacent modules are designed to be parallel when viewed from the side.
This design aims to maximize frictional force under applied wire tension.
However, in such a structure, the constraint of the modules position is only frictional rom the wire, thus it's easy to move the modules when the wire tension is released.
It is not guaranteed that the same contact state and stiffness can be reliably reproduced when applied the tension again.
This is because slack in the wire during tension release is not considered.
In addition, the direction of wire tension is not always perpendicular to the circumferential direction of the contact surface.
Consequently, even when sufficient tension is applied, the effective normal force acting on the contact surface may be reduced, preventing friction from being fully utilized.
The influence of these factors on holding performance has not yet been systematically investigated.

\section{Principle}
\label{sec:principle}

\subsection{Target structures}

As shown in Fig.~\ref{fig:principle_of_wire_jamming}, the wire-based jamming mechanisms have multiple modules, whose geometry affect the properties (e.g., the stiffness and its controllability).
In this study, we investigate multiple mechanisms for finding the suitable design.
However, to realize a wire-based jamming mechanism, several kinematic and structural requirements must be satisfied.
First, in order to maintain the overall shape of the structure, the wire routing must remain fixed.
The structure tends to deform toward a configuration that minimizes the total wire length.
Second, for bending-dominated deformation, the joint axis must be oriented perpendicular to the direction of wire tension.

\begin{figure}
    \centering
    \includegraphics[width=1.0\linewidth]{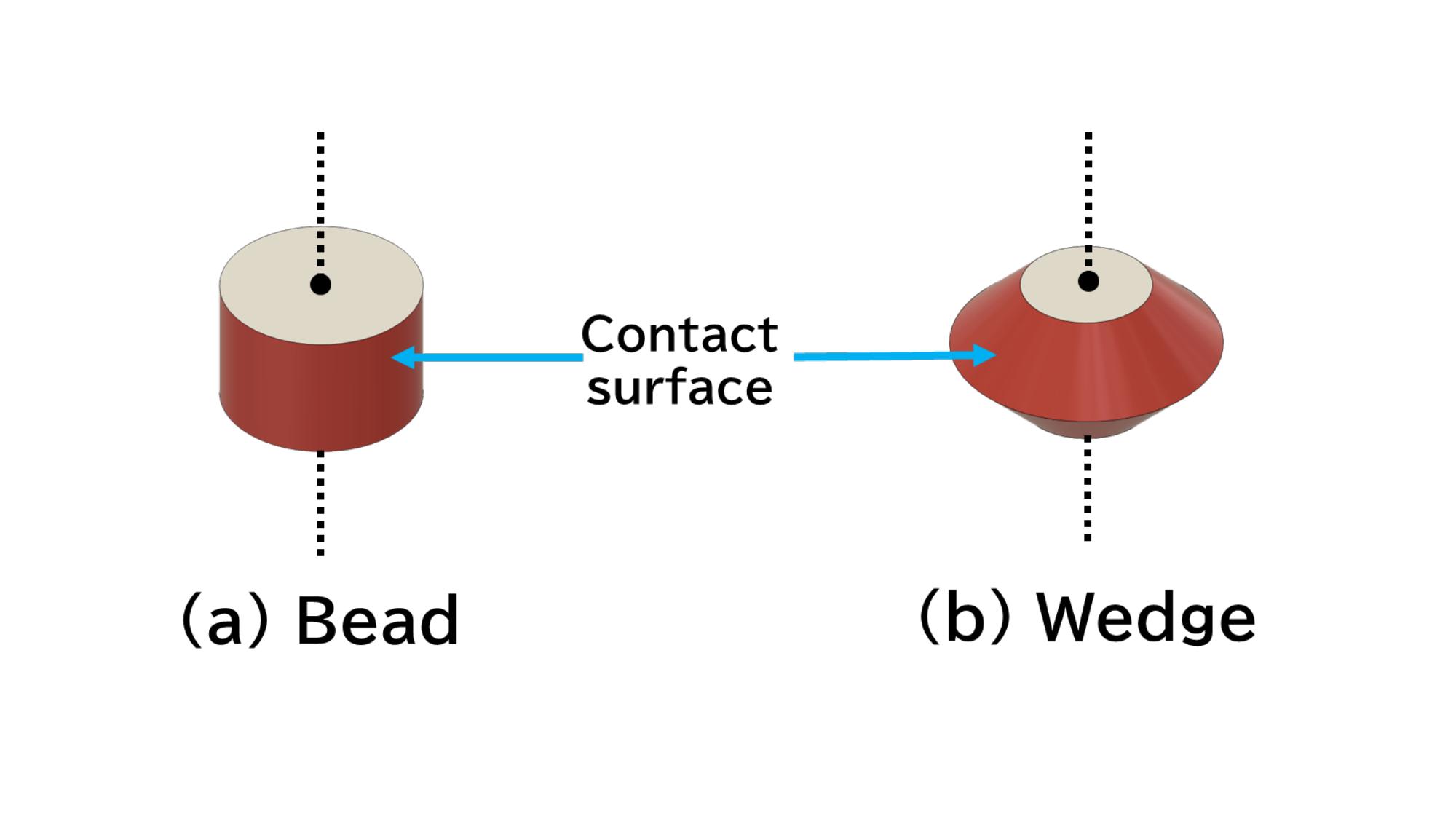}
    \caption{Two types of jamming mechanism}
    \label{fig:concept_of_jamming}
\end{figure}

According to these requirements, we test three types of mechanisms: one is previously proposed design; and others are proposed in this paper with a wedge-shaped contact structure.
Fig.~\ref{fig:concept_of_jamming} illustrates the module geometries considered in this study.
The conventional bead-type jamming mechanism has a single flat contact surface, as indicated by the red regions.
In contrast, the two proposed wedge-based jamming mechanisms introduce inclined contact surfaces, which distribute the force generated by wire tension into both the wire direction and the direction perpendicular to it.
This force redistribution is expected to enhance the effective normal force acting on the contact surfaces and improve holding performance.

\subsection{Bead jamming}
\label{subsec:bead_jamming}
\begin{figure}
    \centering
    \includegraphics[width=1.0\linewidth]{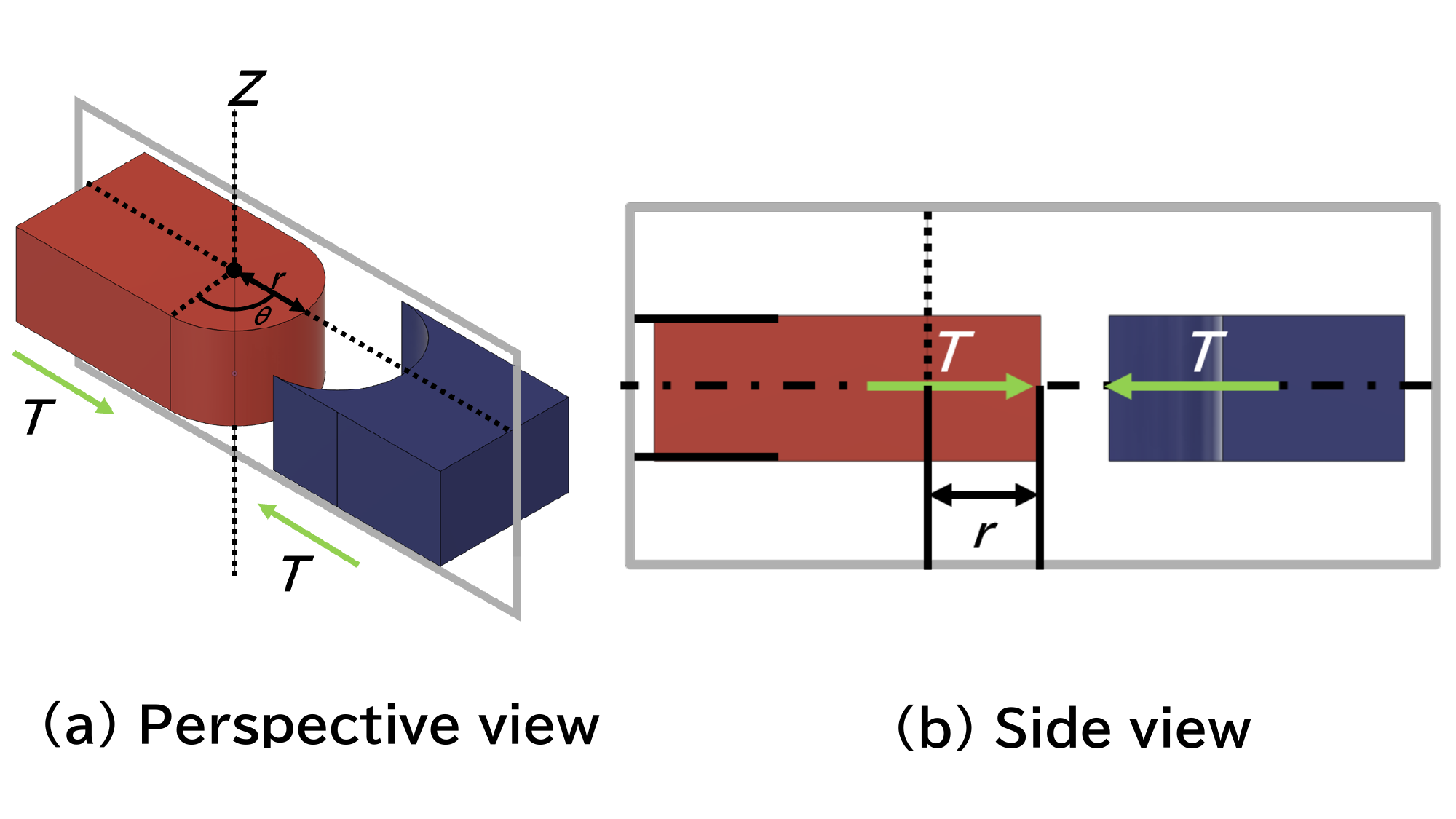}
    \caption{Perspective and side view of bead module}
    \label{fig:Bead_multi_view}
\end{figure}

This shape was proposed for medical applications \cite{sturges1993flexible}\cite{degani2006highly}.
We examine the tension and the force perpendicular to the tension.
The bead jamming mechanism involves only one face of each cylindrical module being in contact.
Fig.~\ref{fig:Bead_multi_view} shows the relationship of the shape of module.
When the wire is pulled, they are forced into contact with each modules, maintaining the wire configuration and preserving its shape.

When the wire is pulled with tension $t$, the normal force $N$ is equal to it, resulting in the holding torque generated between modules by friction, $\tau$, as follows:
\begin{align}
\label{eq:bead_friction}
    \tau = - \mathrm{sign}(\theta) r t
\end{align}
where $\theta$ is the rotation angle around z-axis (a.k.a., joint) and $r$ is the radius of the bead.
That is, the holding torque proportional to $t$ acts as the stiffness that maintains the structure in a straight.

Next, we consider the case where an external force is applied in the joint-axis direction of a bead jamming mechanism.
Fig.~\ref{fig:Joint-axis_exforce} ~(a) illustrates the configuration.
Under the tension $t$, the maximum external force permitted by the equilibrium of forces, $F_d$, is derived.
\begin{align}
\label{eq:bead_break}
    F_d = \frac{H}{2L} t
\end{align}
where $L$ denotes the distance from the joint to the point of application of the external force and $H$ is the height of the module.
Similarly to the above rotational direction, the stiffness proportional to $t$ maintain a straight, although the proportional coefficient differs.

\subsection{Wedge jamming}
\label{subsec:wedge_jamming}
A wedge shape refers to a geometry with a V-shaped or triangular cross-section, characterized by a thin and sharp edge.
It is commonly used to split, fix, or apply pressure to objects by being driven into a gap.
Wedge-shaped structures are widely used in various applications, such as wood joining, firewood splitting, scaffold assembly, and the repair of agricultural tools.

\begin{figure}
    \centering
    \includegraphics[width=1.0\linewidth]{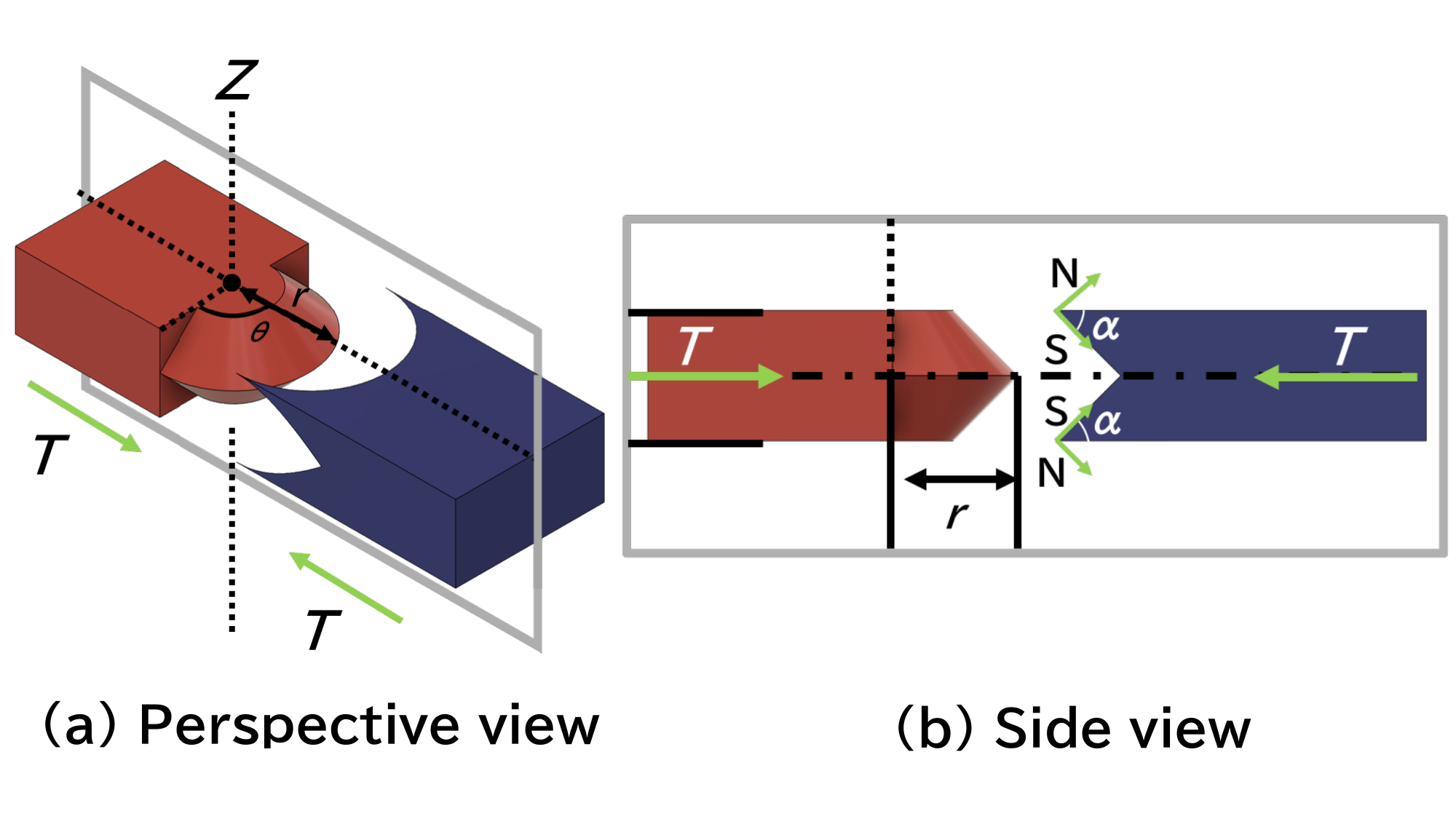}
    \caption{Perspective and side view of wedge module}
    \label{fig:Wedge_multi_view}
\end{figure}

In this study, a wedge-shaped geometry is introduced at the contact interface between adjacent modules.
Fig.~\ref{fig:Wedge_multi_view} illustrates the geometric relationship of the module shape.
When the wire is pulled, the modules are forced into contact with each other, maintaining a fixed wire routing and preserving the overall structure, similar to bead jamming mechanisms.

By introducing the wedge shape with $\alpha$ its angle, this geometry decomposes the tension to two directions: forces perpendicular and parallel to the contact surface, $N$ and $S$, respectively.
In addition, two symmetric contact surfaces are considered.
The tangential force $S$ acts along each wedge surface, whose orientation is defined as $\pm~\alpha$ with respect to the reference axis.
As a result, the vertical components of $S$ have opposite signs and cancel each other, whereas the normal force components act in the same direction and are additive.
Therefore, the contribution of $S$ vanishes in the vertical force balance under the symmetric contact assumption.

\begin{figure}
    \centering
    \includegraphics[width=1\linewidth]{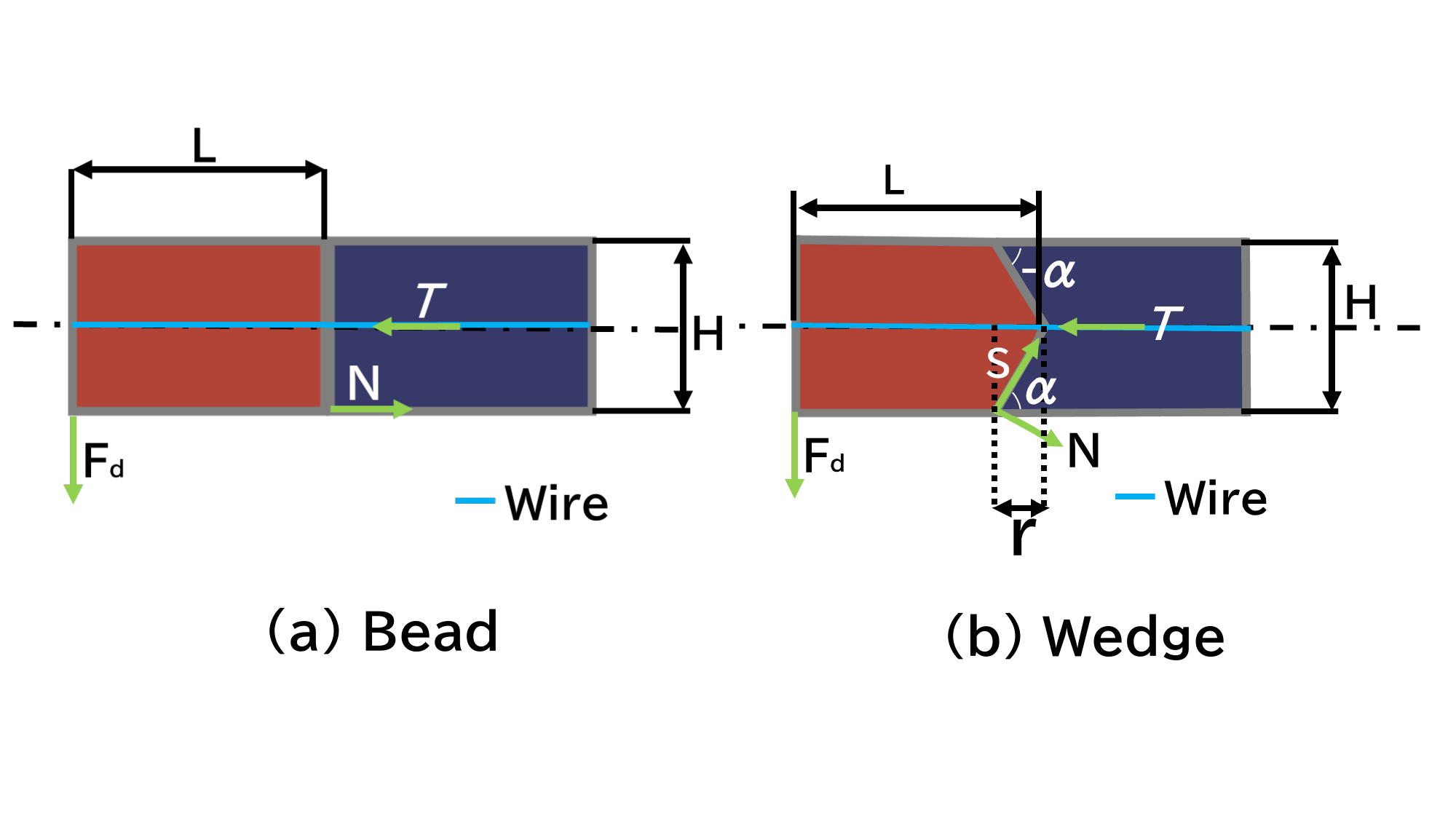}
    \caption{Side view of bead and wedge module}
    \label{fig:Joint-axis_exforce}
\end{figure}

However, with the maximum external force permitted, $F_d$, (see Fig.~\ref{fig:Joint-axis_exforce}(b)), the reaction force to it cancels out the contact force on one side. Consequently, the remaining contact force on the other side is thought to result in the following equilibrium equations.
\begin{align}
\label{eq:wedge_horizontal}
    t + S \cos \alpha + N \sin \alpha = 0
\end{align}
\begin{align}
\label{eq:wedge_vertical}
    F_d = S \sin \alpha - N \cos \alpha
\end{align}
From the equilibrium of moment, the additional can also be derived as well as the case of the bead jamming.
\begin{align}
\label{eq:wedge_moment}
   F_d L = \frac{N H}{2 \sin \alpha}
\end{align}

By solving Eqs.~\eqref{eq:wedge_horizontal},~\eqref{eq:wedge_vertical}, and~\eqref{eq:wedge_moment} simultaneously, $N$ and $S$ are obtained.
\begin{align}
\label{eq:wedge_horizontal_direction}
    N = \frac{2 L \sin^2 \alpha}{2 L \sin \alpha  - H \cos \alpha} t ,
    \quad \left( 0 < \alpha < \pi \right)
\end{align}

\begin{align}
\label{eq:wedge_vertical_distrbance}
    S = \frac{H - 2 L \sin \alpha \cos \alpha}{2 L \sin \alpha - H \cos \alpha} t ,
    \quad \left( 0 < \alpha < \pi \right)
\end{align}

Finally, $F_d$ is derived as follows:
\begin{align}
\label{eq:wedge_force_disturbance}
    F_d = \frac{H}{2 L - H /\tan \alpha} t,
    \quad \left( 0 < \alpha < \pi \right)
\end{align}

In the proposed geometry, the friction between individual modules (i.e., the holding torque) is reduced due to force decomposition from $t$ to $N$ and $S$.
However, $F_d$ becomes larger than the one in the beam jamming with $\alpha < \frac{\pi}{2}$.
That is, the proposed geometry enables adaptable anisotropic force characteristics depending on $\alpha$.

\section{Design of the jamming modules}
Prototypes were developed using the bead and wedge jamming mechanisms, as discussed in Section~\ref{sec:principle} .

\subsection{Jamming module mechanical setup}
In this study, bead jamming is adopted as the baseline mechanism, and two types of wedge-jamming mechanisms are developed: an out-wedge jamming module and an in-wedge jamming module.
All modules are designed based on bead jamming modules with a diameter of 12~mm to ensure a consistent scale.
The rotational range of each joint is set to $\pm \frac{\pi}{6}$.
All modules are fabricated using an FDM-type 3D printer (Bambu Lab H2D) with PLA filament and an infill ratio of 30~\%, to maintain consistent manufacturing conditions. 
Because the modules are fabricated by FDM 3D printing, their contact surfaces possess layer-induced roughness. 
As a result, the frictional behavior is more complex than that assumed in the simplified model. 
The wire is 1.5~mm in diameter, and the wire is terminated using a metal wire clip.

\subsection{Design parameters}
\begin{figure}[tb]
    \centering
    \includegraphics[width=1.0\linewidth]{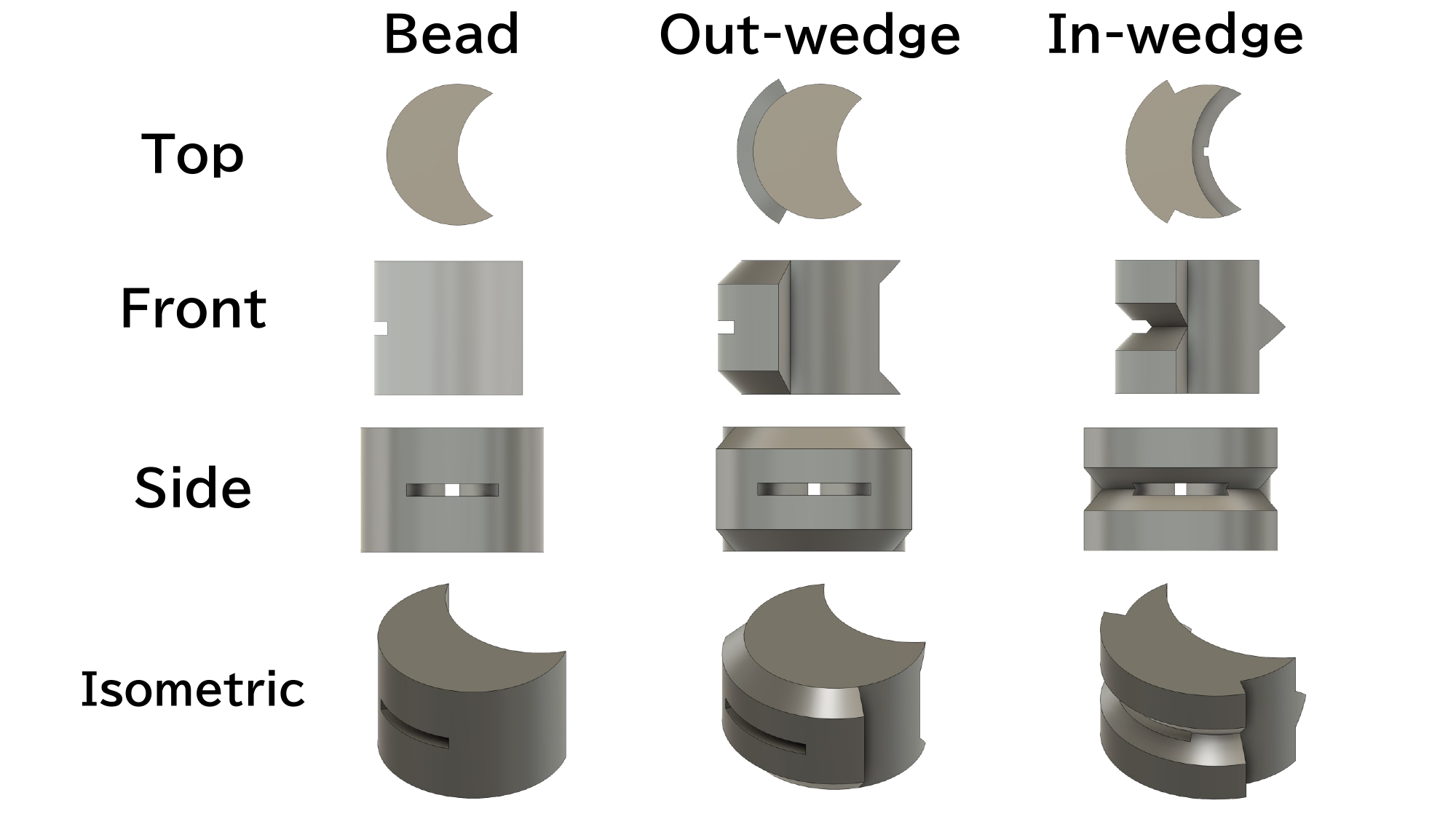}
    \caption{Design illustration of proposed modules}
    \label{fig:design_illustration}
\end{figure}

The wedge-jamming mechanism is designed with parameters that produce two variants: out-wedge and in-wedge jamming.
Fig.~\ref{fig:design_illustration} illustrates the module geometries of the bead jamming, out-wedge jamming, and in-wedge jamming mechanisms.
In the out-wedge jamming module, the wedge angle $\alpha$ is set to $\frac{\pi}{4}$, and wedge structures are placed at both ends of the module.
In the in-wedge jamming module, the wedge angle $\alpha$ is set to $\frac{3}{4}\pi$, and wedge structures are placed at the center of the module.

Between the out-wedge and in-wedge jamming mechanisms, only the parameters proposed in Section~\ref{subsec:wedge_jamming} are varied.
By setting $\alpha = \frac{\pi}{4}$, the analytical result for the maximum external force $F_d$ becomes:
\begin{align}
\label{eq:out-wedge_force_disturbance}
    F_d = \frac{H}{2 L - H} t
\end{align}
From this calculation result, the out-wedge jamming mechanism is expected to improve the holding torque against external forces.

In contrast, by setting $\alpha = \frac{3}{4}\pi$, the analytical result of holding torque against maximum external forces $F_b$  follows:
\begin{align}
\label{eq:in-wedge_force_disturbance}
    F_d = \frac{H}{2 L + H} t
\end{align}
From this calculation result, the in-wedge jamming mechanism is expected to reduce the holding torque against external forces.
Since the in-wedge structure is located near the center of the module, the moment arm becomes very small.
Therefore, the holding torque against external torsion is expected to improve.

The effects of in-wedge jamming and out-wedge jamming are verified experimentally in the next section.

\section{Experiment}
In this section, the performance of the proposed method is experimentally validated.
Three experiments were conducted to evaluate the holding torque characteristics of the jamming mechanisms.
The holding torque was measured using a drill stand as vertical guide and two digital force gauges with different measurement ranges (GOYOJO Digital Force Gauge: 0.001-5~N·m and 0.1-500~N·m).

A series elastic actuator \cite{rollinson2013design} (Hebi Robotics: X5-9, peak torque 13~Nm) and a 3D-printed pulley were used to generate the tensile force in the wire.
The series elastic actuator can measure the joint-axis torque directly.
The wire tension was calculated from the pulley radius and the measured motor torque, enabling both control of the wire tension and evaluation of the jamming performance.

\begin{figure}[tb]
    \centering
    \includegraphics[width=1.0\linewidth]{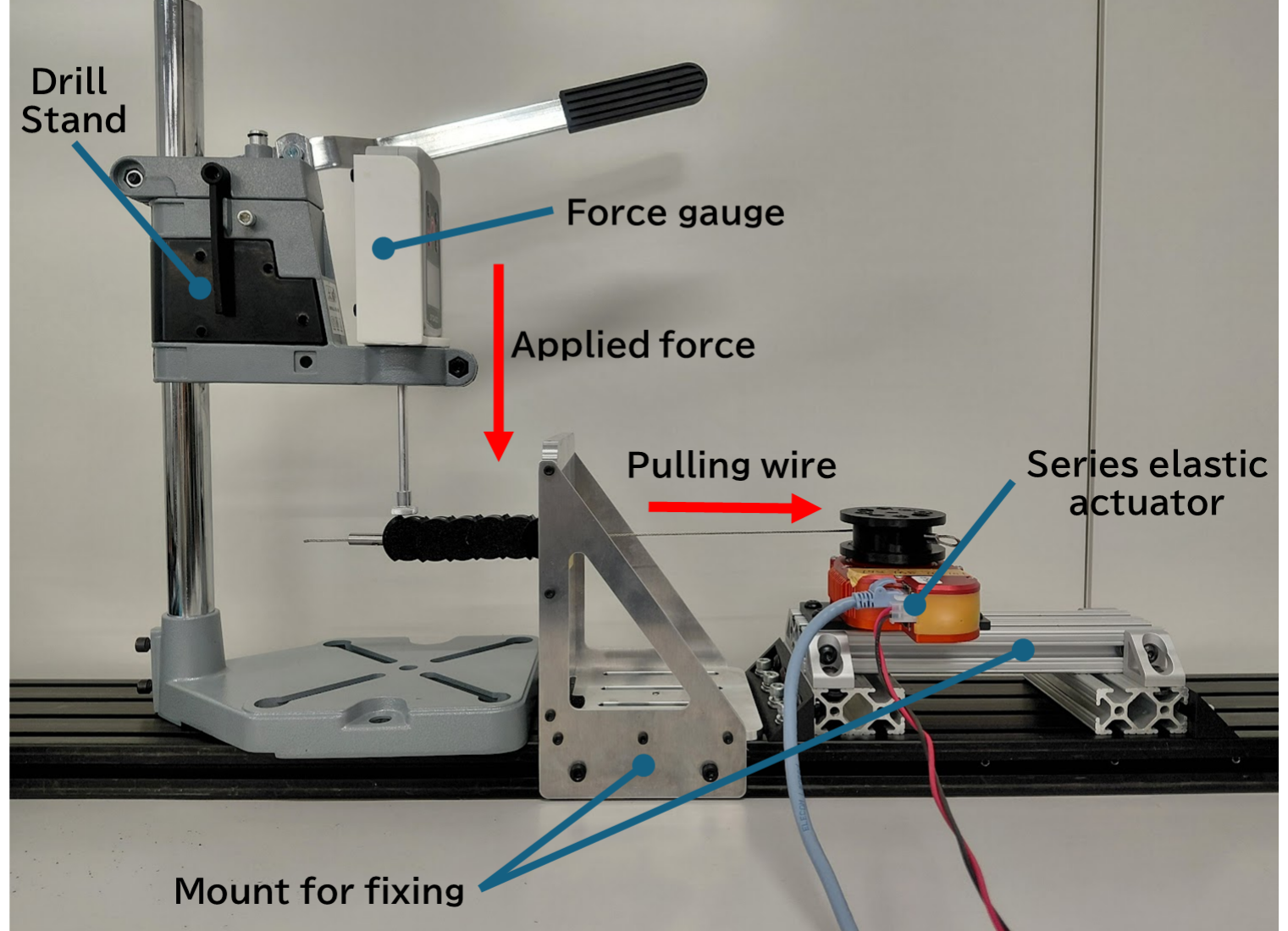}
    \caption{Force test bench used to evaluate the holding torque of the wire-based jamming mechanism}
    \label{fig:test_bench_force}
\end{figure}

\begin{figure}[tb]
    \centering
    \includegraphics[width=1.0\linewidth]{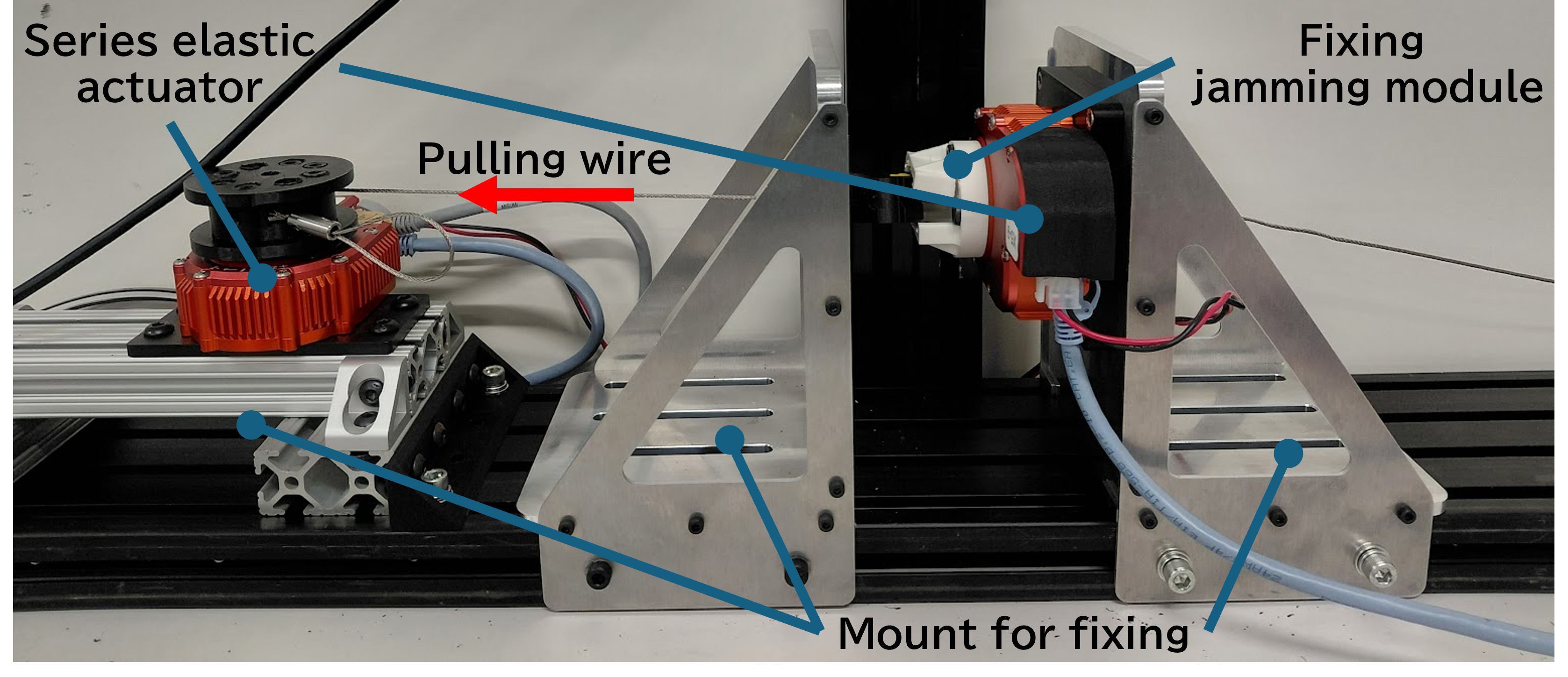}
    \caption{Torsion test bench used to evaluate the torsional holding torque of the wire-based jamming mechanism. The series elastic actuator applies a torsional torque to the module, and the angular displacement is measured using the motor encoder.}
    \label{fig:test_bench_momentam}
\end{figure}

To evaluate the proposed method, two types of test benches were prepared, as shown in Figs.~\ref{fig:test_bench_force} and~\ref{fig:test_bench_momentam}.

\subsection{Relationship between wire tension and average holding torque}
\label{sec:exp1}
\subsubsection{Measurement method}
\begin{figure}
    \centering
    \includegraphics[width=1.0\linewidth]{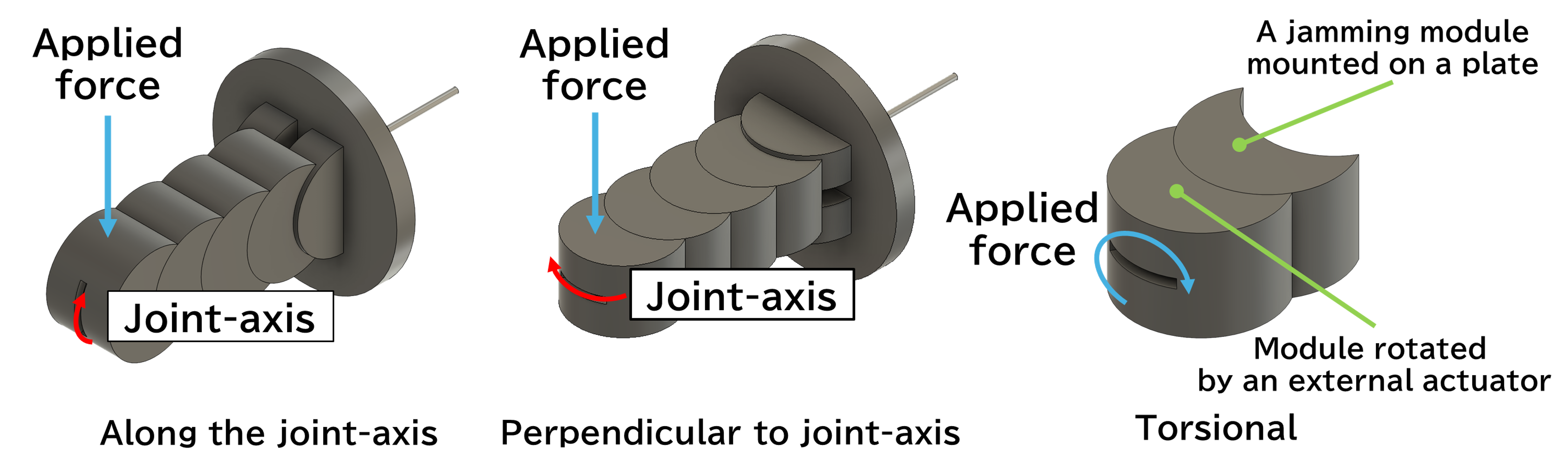}
    \caption{The image of three types of disturbance directions: joint-axis direction, perpendicular to joint-axis direction and torsional}
    \label{fig:disturbance_direction}
\end{figure}

To evaluate the relationship between wire tension and average holding torque, the force test bench shown in Fig.~\ref{fig:test_bench_force} was used.
The experimental procedure was as follows:

\begin{enumerate}[a)]
    \item Five jamming modules were connected in series using a wire. When tension was applied to the wire by the motor, the jamming phenomenon occurred in the modules.

    \item The wire tension was calculated from the motor torque and the pulley radius. The applied tension was set to 10~N, 20~N, 30~N, 40~N, and 50~N.

    \item While maintaining the specified wire tension, an external force was applied to the tip of the module chain using the vertical guide.

    \item The external force was gradually increased until the base of the modules detached. The peak torque at this moment was recorded as the maximum holding torque.

    \item For each tension level, the experiment was repeated 25 times. As shown in Fig.~\ref{fig:disturbance_direction}, measurements were conducted for disturbances applied along the joint-axis direction and perpendicular to joint-axis direction, with the direction adjusted to upward and downward by rotating the setup. The average maximum holding torque was then calculated.
\end{enumerate}

\subsubsection{Evaluation and discussion}
\begin{figure}[tb]
    \centering
    \includegraphics[width=1.0\linewidth]{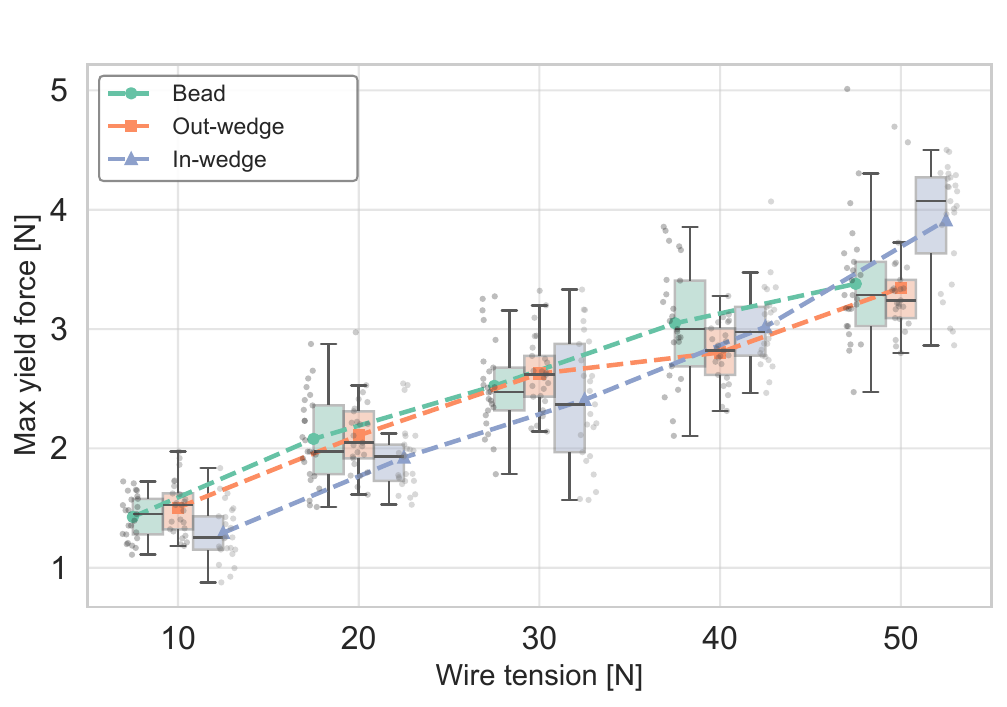}
    \caption{Relationship between wire tension and holding torque in the joint-axis direction disturbance}
    \label{fig:exp1_normal_result}
\end{figure}

\begin{figure}[tb]
    \centering
    \includegraphics[width=1.0\linewidth]{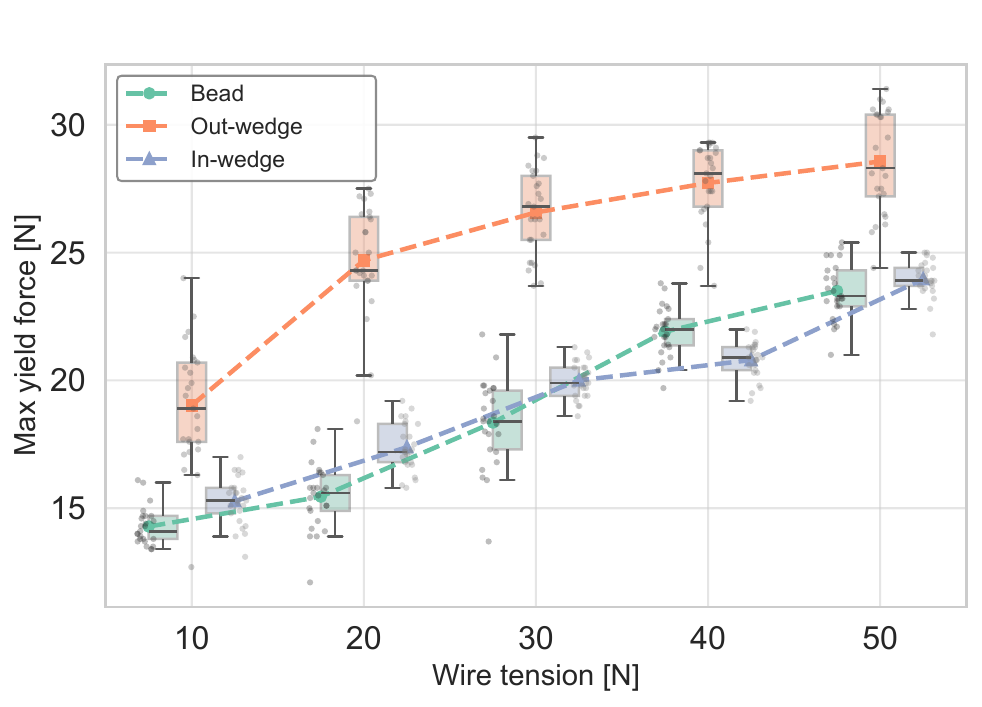}
    \caption{Relationship between wire tension and holding torque in the perpendicular to joint-axis direction disturbance}
    \label{fig:exp1_vertical_result}
\end{figure}

The experimental results for the joint-axis direction and perpendicular to joint-axis direction disturbance directions are shown in Figs.~\ref{fig:exp1_normal_result} and~\ref{fig:exp1_vertical_result}, respectively.
Each figure presents the results of 25 trials as box plots, and the mean values are connected by dashed lines.

As shown in Fig.~\ref{fig:exp1_normal_result}, the holding torque in the joint-axis direction generally increases with increasing wire tension.
In the proposed mechanism, the holding torque in the joint-axis direction is redistributed into two directional components due to the wedge structure.
Therefore, it is expected that the bead configuration exhibits the largest holding torque in this direction.
However, the holding torque of the out-wedge configuration shows no significant difference from that of the bead configuration.
This result indicates that introducing the wedge structure maintains a comparable level of holding torque in the joint-axis direction.

Next, the holding torque in the perpendicular to joint-axis direction is considered, as shown in Fig.~\ref{fig:exp1_vertical_result}.
The out-wedge configuration exhibits the largest holding torque in the perpendicular to joint-axis direction.
This result indicates that setting the wedge angle $\alpha = \frac{\pi}{4}$ enables the wire tension to be effectively converted into a holding force in the joint-axis direction.

\subsection{Relationship between the number of joints and average holding torque}
\subsubsection{Measurement method}
\begin{figure}
    \centering
    \includegraphics[width=1.0\linewidth]{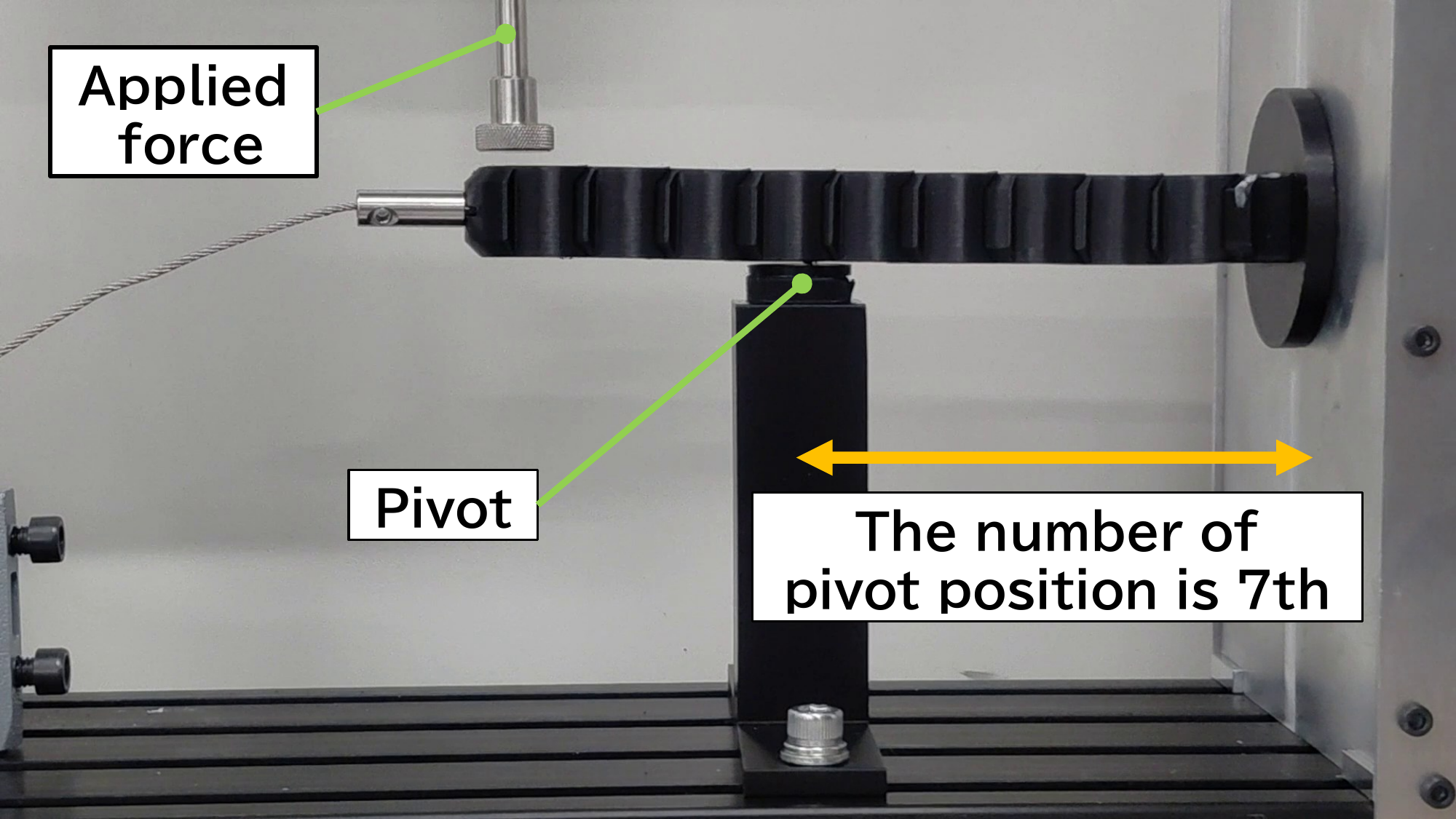}
    \caption{The experiment environment of relationship between the number of joints and average holding torque. In this image, it shows the example of pivot position 7th.}
    \label{fig:pivot_point}
\end{figure}

To evaluate the influence of the pivot position on the holding torque, the same force test bench shown in Fig.~\ref{fig:test_bench_force} was used.
The experimental procedure was identical to that of Experiment~1, except for the following changes:

\begin{enumerate}[a)]
    \item The wire tension was fixed at 30~N and ten modules were connected in series.
    \item The pivot position of the module chain was varied among the 3rd, 5th, and 7th modules from the base plate as show in Fig.~\ref{fig:pivot_point}, with disturbances applied at the distal end. As the pivot index increased, the distance between the support point and the point of force application decreased.
\end{enumerate}

\subsubsection{Evaluation and discussion}
\begin{figure}[tb]
    \centering
    \includegraphics[width=1.0\linewidth]{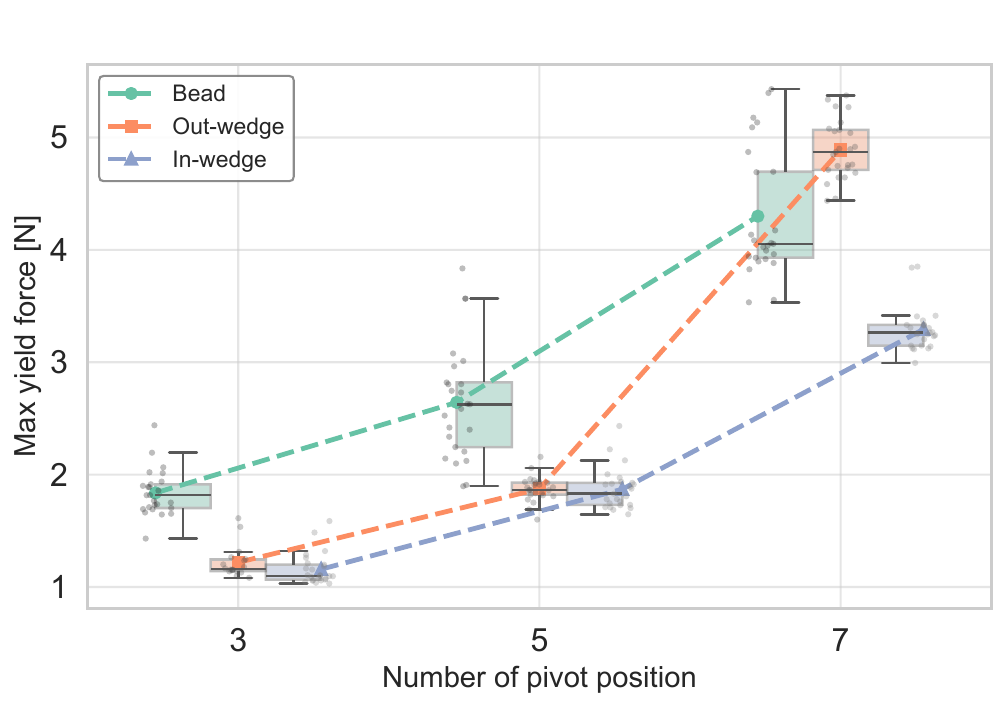}
    \caption{Holding torque in the joint-axis direction disturbance direction with different pivot position}
    \label{fig:exp2_pivot_point_normal_result}
\end{figure}

\begin{figure}[tb]
    \centering
    \includegraphics[width=1.0\linewidth]{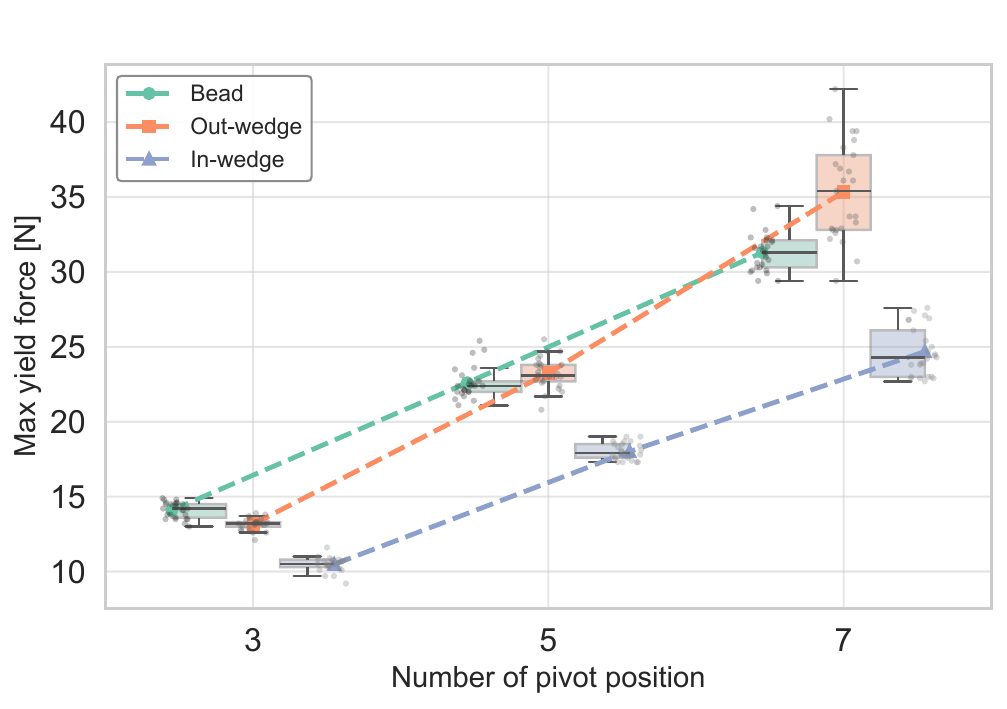}
    \caption{Holding torque in the perpendicular to joint-axis disturbance direction with different pivot position}
    \label{fig:exp2_pivot_point_vertical_result}
\end{figure}

The experimental results are shown in Figs.~\ref{fig:exp2_pivot_point_normal_result} and~\ref{fig:exp2_pivot_point_vertical_result}.
As in Experiment~1, each figure presents the results of 25 trials as box plots, and the mean values are connected by dashed lines.

As shown in Fig.~\ref{fig:exp2_pivot_point_normal_result}, the holding torque in the joint-axis direction generally increases as the distance between the support point and the point where the external force is applied decreases.
Similar to the results of Experiment~1, the bead configuration exhibits the largest holding torque in the joint-axis direction.

However, when the support point is located at the 7th module, the out-wedge configuration exhibits the highest holding torque.
This occurs because the proximity of the support point to the disturbance point suppresses the relative displacement between modules.
Therefore, when the number of modules is small, the out-wedge configuration becomes effective.

A similar tendency is observed for the holding torque in the perpendicular to joint-axis direction as show in Fig.~\ref{fig:exp2_pivot_point_vertical_result}.
When the support point is located at the 7th module, the out-wedge configuration exhibits larger holding torque.
This result is consistent with the findings in Experiment~1, indicating that the out-wedge module is effective against disturbances in the perpendicular to joint-axis direction.

\begin{figure*}
    \centering
    \includegraphics[width=1.0\linewidth]{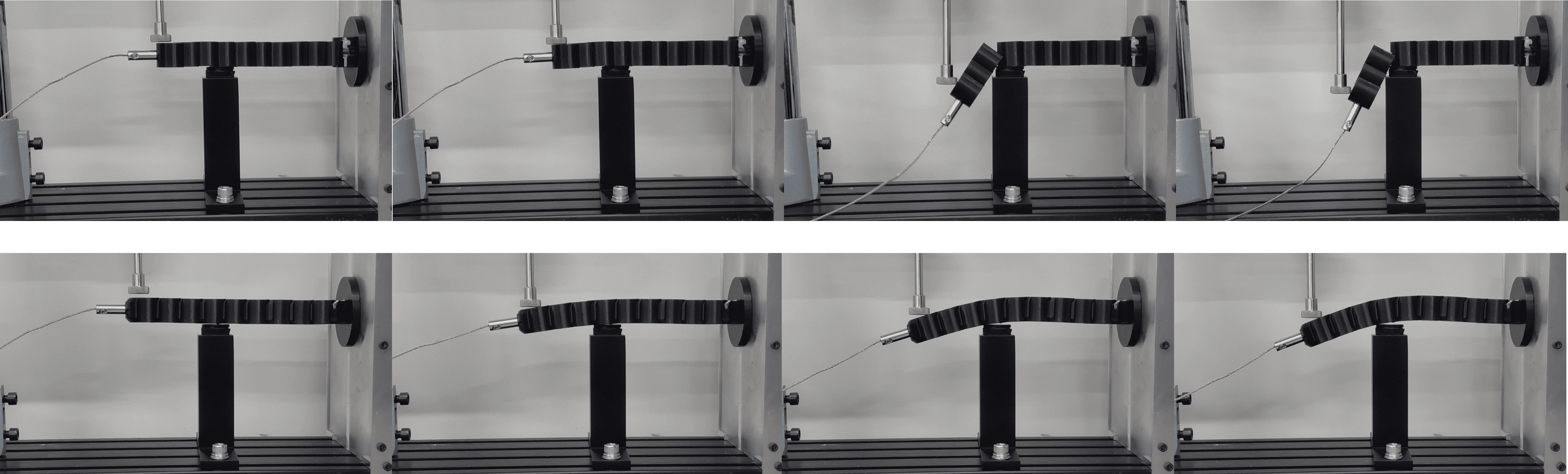}
    \caption{The figure shows the behavior of bead and out-wedge modules in slow motion when a perpendicular to joint-axis disturbance is applied with the pivot position set at the 7th from the base plate. The upper figure shows the bead modules, and the lower figure shows the out-wedge modules.}
    \label{fig:exp2_jamming_change}
\end{figure*}

However, the difference compared to the bead configuration is smaller than expected during the design stage.
One possible reason is that the perpendicular to joint-axis holding torque of the out-wedge structure becomes excessively large.
When a force is applied at the tip, the modules between the support point and the base can lift upward, which weakens the constraint between modules (see Fig.~\ref{fig:exp2_jamming_change}).

\subsection{Relationship between wire tension and torsional holding torque}
\subsubsection{Measurement method}
The torsional holding performance of the jamming mechanism was evaluated using the test system shown in Fig.~\ref{fig:test_bench_momentam}.
Two modules were connected, and while applying a specified wire tension, a torsional torque was applied using another motor.

The torsional resistance was evaluated by measuring the angular displacement obtained from the motor encoder.
The wire tension was varied from 0~N to 40~N in increments of 10~N.
At each tension level, a torsional torque was applied in increments of 0.1~Nm every second until reaching 1~Nm.

\subsubsection{Evaluation and discussion}
\begin{figure}[tb]
    \centering
    \includegraphics[width=1.0\linewidth]{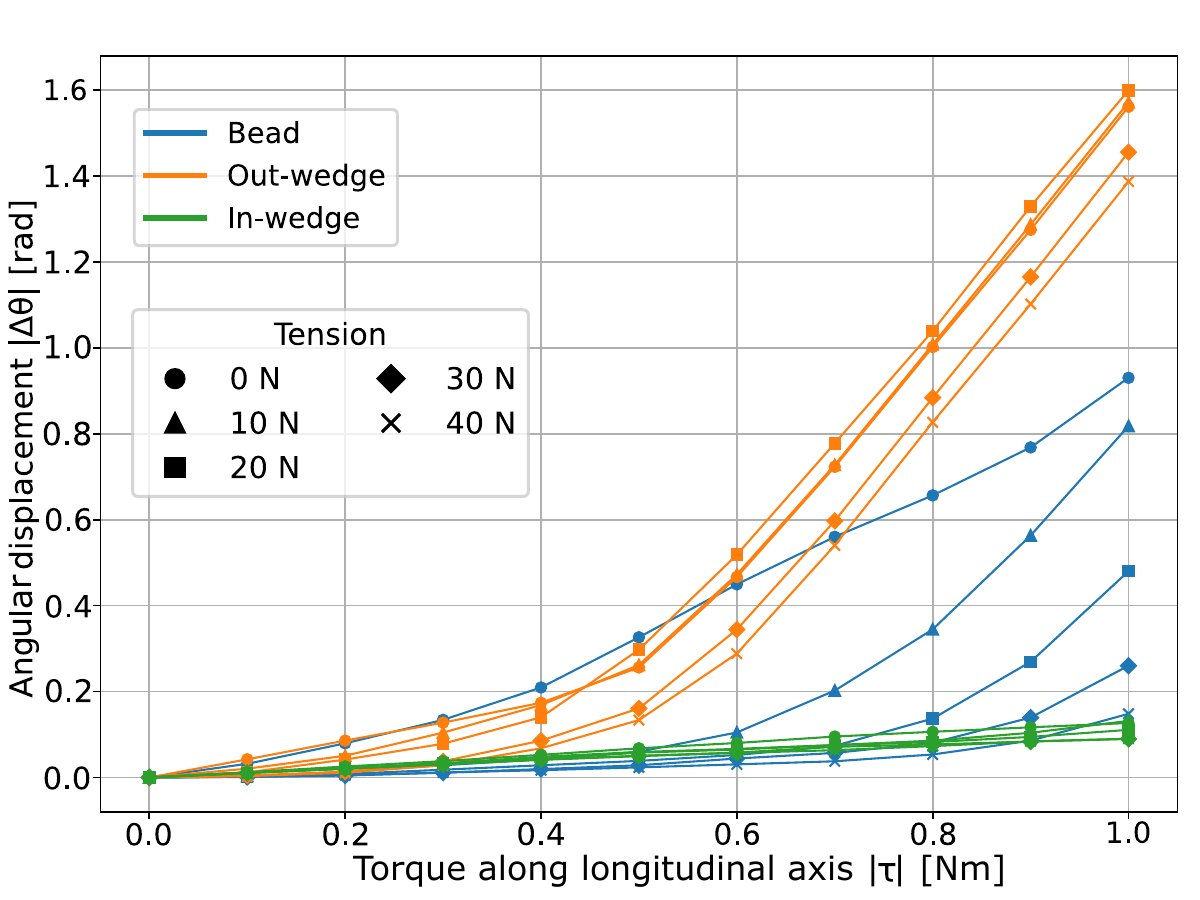}
    \caption{Relationship between torsional torque, angular displacement, and wire tension}
    \label{fig:exp3_torsion_result}
\end{figure}

Fig.~\ref{fig:exp3_torsion_result} shows the relationship between the wire tension and the encoder displacement, where the vertical axis represents the wire tension and the horizontal axis represents the encoder change.
In all configurations, the angular displacement in the twisting direction decreases as the wire tension increases.
This indicates that increasing the wire tension enhances the jamming effect, thereby suppressing torsional deformation.

Among the tested configurations, the out-wedge structure exhibits the smallest holding torque against torsional disturbance.
This occurs because the inclination direction of the out-wedge surface aligns with the direction of the applied torsional torque, preventing the wire tension from being effectively converted into a force that suppresses twisting.
In addition, the distance between the torsional rotation center and the wedge structure is relatively large, resulting in a longer moment arm that reduces the effectiveness of the applied force in resisting torsional deformation.

In contrast, the in-wedge structure exhibits a large holding torque even at 0~N of wire tension.
This is because the orientation of the in-wedge surface is nearly orthogonal to the direction of the applied torsional torque, allowing the structure to effectively resist twisting forces.

Furthermore, the distance between the torsional rotation center and the wedge structure is relatively small.
As a result, the moment arm becomes smaller, allowing the applied force to efficiently generate resistance against torsional deformation.

\section{Conclusion}
In this study, we proposed a wedge-jamming mechanism for wire-based jamming structures.
By introducing wedge-shaped contact surfaces into the modules, the ratio of horizontal and vertical force components generated by wire tension can be controlled by the wedge angle.

Two types of wedge-based modules, namely the out-wedge and the in-wedge configurations, were fabricated and experimentally evaluated. 
The results showed that the out-wedge configuration is most effective against disturbances in the perpendicular to joint-axis direction, whereas the in-wedge configuration provides superior resistance to torsional disturbances. 
These findings suggest that, in wedge-jamming mechanisms, a wedge angle satisfying $(0 < \alpha < \frac{\pi}{2})$ effectively enhances the holding capability against vertical disturbances, while a shorter distance from the wedge center improves the holding capability against torsional disturbances.

While maintaining the trade-off relationship with the joint-axis holding torque, which has been the main focus of previous studies, the proposed mechanism enables the holding characteristics against vertical and torsional disturbances to be tuned simply by changing the wedge angle.
These results expand the applicability of wire-based jamming mechanisms by enabling directional control of holding performance.

In future work, we will investigate the optimization of the wedge geometry and the integration of the proposed mechanism into robot limbs, such as the feet and arms, where contact with the environment frequently occurs.
\balance

\bibliographystyle{IEEEtran}
\bibliography{IEEEabrv,biblio_traps_dynamics}

\end{document}